\documentclass[runningheads]{llncs}

\usepackage{eccv}

\usepackage{eccvabbrv}

\usepackage{graphicx}
\usepackage{booktabs}

\usepackage[accsupp]{axessibility}  

\usepackage[pagebackref,breaklinks,colorlinks,citecolor=eccvblue]{hyperref}

\usepackage{orcidlink}

\usepackage{colortbl}
\usepackage{xcolor}
\usepackage{subcaption}
\usepackage{graphicx}
\usepackage{multirow}
\usepackage{lipsum}  
\usepackage{booktabs}
\usepackage{duckuments}
\usepackage{seqsplit}

\usepackage{caption}
\usepackage{subcaption}

\newcommand{\Real}{\mathbb{R}}

\newcommand{\bK}{\mathbf{K}}

\newcommand{\gray}[1]{\textcolor{gray}{#1}}

\newcommand{\conditionalcomment}[1]{\if\commenttext1 \else {#1} \fi}
\newcommand{\grayconditionalcomment}[1]{\if\commenttext1 \else \gray{{#1}} \fi}

\definecolor{grey}{rgb}{0.9, 0.9, 0.9}

\usepackage{amsmath,amsfonts,bm}

\def\Tableref#1{Table~\ref{#1}}

\def\Figref#1{Figure~\ref{#1}}

\def\Secref#1{Section~\ref{#1}}

\def\eqref#1{equation~\ref{#1}}
\def\Eqref#1{Equation~\ref{#1}}

\def\1{\bm{1}}

\def\vc{{\bm{c}}}

\def\vf{{\bm{f}}}

\DeclareMathAlphabet{\mathsfit}{\encodingdefault}{\sfdefault}{m}{sl}
\SetMathAlphabet{\mathsfit}{bold}{\encodingdefault}{\sfdefault}{bx}{n}

\begin{document}

\title{
In Defense of Lazy Visual Grounding
\\ for Open-Vocabulary Semantic Segmentation}

\titlerunning{Lazy Visual Grounding}

\author{Dahyun Kang \quad Minsu Cho\vspace{0.15cm}
}

\authorrunning{Kang and Cho}

\institute{Pohang University of Science and Technology (POSTECH), South Korea\\
{
\small
\url{https://cvlab.postech.ac.kr/research/lazygrounding}
}}

\maketitle

\begin{abstract}
We present lazy visual grounding, a two-stage approach of unsupervised object mask discovery followed by object grounding, for open-\seqsplit{vocabulary} semantic segmentation.
Plenty of the previous art casts this task as pixel-to-text classification without object-level comprehension, leveraging the image-to-text classification capability of pretrained vision-and-language models.
We argue that visual objects are distinguishable without the prior text information as segmentation is essentially a vision task.
Lazy visual grounding first discovers object masks covering an image with iterative Normalized cuts and then later assigns text on the discovered objects in a late interaction manner.
Our model requires no additional training yet shows great performance on five public datasets: 
Pascal VOC, Pascal Context, COCO-object, COCO-stuff, and ADE 20K.
Especially, the visually appealing segmentation results demonstrate the model capability to localize objects precisely.


\end{abstract}


\vspace{-10mm}

\begin{figure*}[!ht]
	\centering
	\small
    \includegraphics[width=\linewidth]{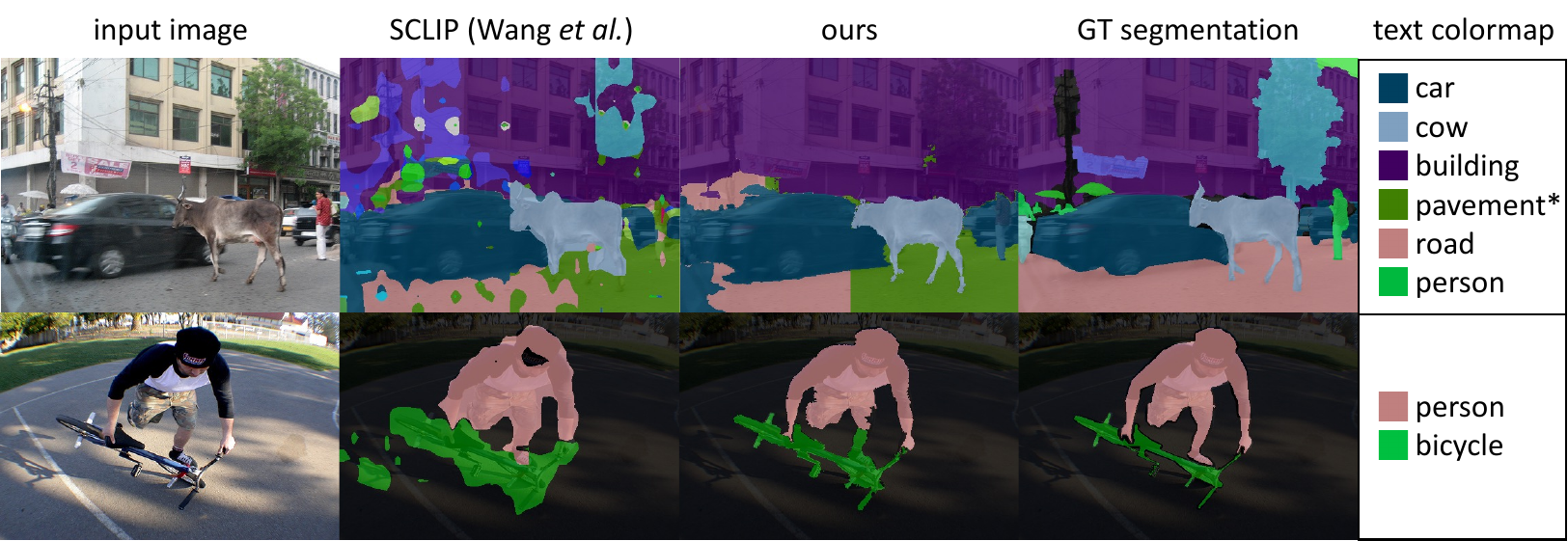}
 \caption{\textbf{Comparison of a pixel-grounding method and our object-grounding method.}
 A pixel-grounding method (SCLIP~\cite{sclip}) often produces spurious correlations with imprecise edges.
 Unlike this pixel-to-text classification approach for open-vocabulary segmentation,
 we solve this task with lazy visual grounding: a two-stage approach of \textit{object mask discovery} followed by \textit{object grounding} in a late interaction manner.
 }
\label{fig:teaser}
\vspace{-5mm}
\end{figure*}

\section{Introduction}
Image segmentation is a fundamental task in computer vision, aiming to partition an image into multiple segments with coherent characteristics~\cite {thoma2016survey, li2018survey}.
The classic approach to the problem is to collect nearby image pixels with shared characteristics using graph partitioning~\cite{normalizedcut, blake2011markov, boykov2001fast}, feature clustering~\cite{wu1993optimal, chen1998image}, histogram equalization~\cite{pizer1987adaptive}, edge detection~\cite{kass1988snakes, arbelaez2009contours}.
Deep neural networks for image segmentation~\cite{fcn, deeplab, unet, deconvnet, maskformer} have relied on supervised training with segmentation datasets of pixel-level class supervision~\cite{pascal, coco, sam, lvis, ade20k, cityscapes}.
However, such models are trained with the fixed set of classes hence unable to recognize unseen classes beyond the predefined set of seen classes.

The task of open-vocabulary semantic segmentation (OVSeg)~\cite{z3net, spnet, groupvit, lseg, openseg} is an image segmentation task of which classes are grounded with text.
OVSeg aims to align image segments with potentially unseen classes described in text, spanning the modality of classic image segmentation to multi-modal.
To solve this task, image-text encoders such as CLIP~\cite{clip} are often used to recognize the text descriptions.
A line of work~\cite{sclip, zeroseg, tcl, maskclip} leverages its alignment of image patch and text embeddings to solve OVSeg by pixel-to-text classification.
However, this pixel grounding approach hardly comprehends distingushable objects\footnote{
This is probably because CLIP is trained to represent the holistic image through a separate CLS token embedding, and the patch embeddings are not explicitly trained to represent the corresponding image patch.}~(\textit{cf.} Fig.~\ref{fig:teaser}).
The imprecise segmentation of the modern multi-modal segmentation models encourage us to revisit the classic computer vision techniques -- regarding an image with a bag of segmentable visual units.
Imagine a four-legged animal on grass that we want to segment out.
Do we need its name before segmentation?
Having been approached since decades ago, image segmentation is \textit{primarily a vision task}, and we argue that the objects can be distinguished as visual units given the vision input only without knowing their names in text.  

Our approach, dubbed \textbf{La}zy \textbf{V}isual \textbf{G}rounding (LaVG), first partitions an image into image segments using a classical image segmentation method, which is based on the vision input and no textual priors on the target classes.
The text entity is later assigned to the partitioned segments.
For image segmentation, we revisit the related task of unsupervised object discovery~\cite{lost, phm, vo2019unsupervised, vo2020toward} which aims to distinguish visual units in bounding boxes or object masks without semantic priors such as classes.
Following the recent unsupervised object discovery methods~\cite{cutler, most, tokencut}, we retrocede the Normalized Cut~\cite{normalizedcut} algorithm.
Normalized cut bipartitions a weighted graph into two sets of nodes such that the affinity between the bipartition is minimal.
Following the recent object discovery approach~\cite{cutler, tokencut, instancecut, linearspectralclustering}, we apply Normalized cut on the outputs of a self-supervised vision transformer, DINO~\cite{dino}, which present an empirical property that attends to the salient object region.
We iteratively apply Normalized cut until the identified segments cover all image pixels for semantic segmentation, \ie, panoptic segmentation~\cite{panopticsegmentation}, and name this approach as \textit{Panoptic cut}.
Once Panoptic cut is finished, then the objects are grounded with text.
Each discovered segment is aggregated to form an object prototype embedding and finally matched with given text embedding from a free-form text (\Figref{fig:imagesinthewild}).
Note that the this two-stage process requires \textit{no additional training} hence significantly reduces computational cost.
LaVG is simple and training-free yet has shown its superiority over not only the training-free couterpart models~\cite{sclip, clippy, zeroseg, maskclip, reco} but also the trained models~\cite{vilseg, groupvit, segclip, viewco, mixreorg, ovsegmentor, tcl} that used image-text paired datasets~\cite{cc3m, yfcc, coco}.
It is noteworthy that our model localizes objects qualitatively more precisely than the pixel-to-text classification methods (\Figref{fig:teaser}).

Our contributions can be summarized as follows:

\begin{figure*}[t!]
    \centering
    \small
    \includegraphics[width=\linewidth]{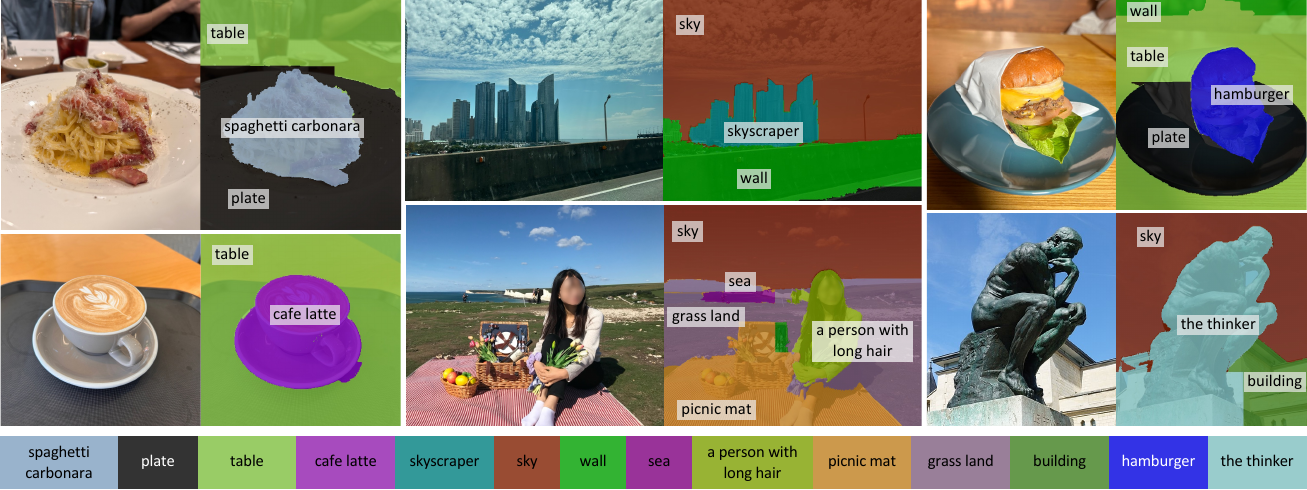}
 \caption{Segmentation results of LaVG given the text description set. These precise object masks and visual grounding results are produced \textit{in a training-free fashion}.
 }
\label{fig:imagesinthewild}
\end{figure*}

\begin{itemize}
    \item We present Lazy Visual Grounding (LaVG) which \textit{discovers objects first and then later assigns text on the objects} in a late interaction manner for open-vocabulary semantic segmentation.
    \item We revisit Normalized cut for iterative discovery of object masks until the masks cover all the image pixels, which we dub as Panoptic cut.
    \item Our model requires no additional training and thus introduces zero training computation cost.
    \item Our simple method based on graph partitioning quantitatively surpasses the state-of-the-art open-vocabulary segmentation models and produces segmentation masks with remarkably more precise object boundaries.
\end{itemize}


\section{Related Work}
\subsection{Unsupervised Object Discovery}
Object discovery~\cite{tuytelaars2010unsupervised} has been a fundamental computer vision task.
In 2000s, unsupervised object discovery started with discovering the image groups containing objects of the same class with analytic techniques~\cite{russell2006using, sivic2005discovering, weber2000towards, grauman2006unsupervised}.
Later, it advances to identify the location of the objects~\cite{zhu2014unsupervised, zhang2015mining, lee2009shape}, which is viewed as the seminal work of the object discovery task with object localization.
Object discovery has been formulated with an input of an image collection~\cite{phm, vo2019unsupervised, vo2020toward}, 
where objects are discovered among the image collection via an image pair matching.
Recently, a self-supervised vision transformer named DINO~\cite{dino} has shown the emerging property of precisely attending the foreground objects within the transformer~\cite{transformers} attention layer.
As it provides a strong clue for foreground object saliency, DINO completely revamps the paradigm of unsupervised object discovery~\cite{simeoni2023unsupervisedsurvey, dinosour, selfmask, freesolo}.
Pioneered by LOST~\cite{lost}, the predominating approach for object discovery is to distinguish existing objects based on the DINO activation map.
One approach is seed finding~\cite{lost, found}, which explores foreground/background region from a highly confident ``seed'' pixel on the attention map.
The other popular approach leverages the graph-partitioning segmentation, \eg, Normalized cut~\cite{normalizedcut}, on the DINO attention map to segregate the foreground objects from the background~\cite{tokencut, cutler, most, locate}, often followed by self-training of instance segmentation models~\cite{maskrcnn, cascadercnn} by the obtained masks~\cite{most, cutler, locate}.
Our work is motivated by the second graph-partitioning approach, which robustly identifies foreground objects.

\subsection{Self-Supervised Vision Transformers}
Deep self-supervised learning models~\cite{jing2020self, gui2023survey} are trained with the self-supervision arisen from the input data rather than external human-annotated labels.
The self-supervised learning paradigm has shown effective in terms of task performance as well as generalization ability across multiple downstream tasks~\cite{wu2018unsupervised, deepcluster, zhuang2019local, simclr, swav, dino, cms} without task-specific annotations.
One popular approach~\cite{dino, dinov2, grill2020bootstrap} trains a model to map an image and its another view to nearby points on the feature space.
This class-agnostic objective encourages the image encoder to capture visual invariancy.
Unlike class-supervised models that often attend on the specific part that represents an object, a representative self-supervised model named DINO~\cite{dino} shows the emerged localization ability capturing foreground objects on the attention~\cite{transformers}.
This property has benefited several dense prediction tasks such as weakly-supervised few-shot segmentation~\cite{cst} and unsupervised image/video segmentation~\cite{ziegler2022self, zadaianchuk2022unsupervised, yin2022transfgu}.
In this work, we leverage these foreground-attended representations as an input source to Normalized cut.

\subsection{Vision-and-Language Representation Learning}
Recent advances in efficient and modern hardware accelerate training huge foundation models at scale~\cite{florence, llava, gpt3, coca, lit}.
Among them, a representative vision and language model named CLIP~\cite{clip} is trained using 400 million web-crawled image and text pairs to learn cross-modal embedding similarity.
CLIP trains two-tower image-text transformer encoders such that embedding similarity between a positive image-text pair is larger than other pairs in a training batch.
The open-sourced CLIP models are leveraged to a number of vision and language work including zero-shot classification~\cite{liu2023learning, iclip}, long-tailed recognition~\cite{vlltr, ma2022unleashing, ma2021simple}, referring segmentation~\cite{risclip, wang2022cris}, cross-modal retrieval~\cite{Huang_2023_CVPR, ma2022ei} as well as OVSeg.

\subsection{Open-Vocabulary Semantic Segmentation (OVSeg)}
Open-vocabulary segmentation is a pixel-level image classification task with categories described in text.
Compared to the classic definition of semantic segmentation~\cite{fcn, deeplab} with a closed set of predefined target classes, the target class set of OVSeg consists of \textit{open-vocabulary text} and is not necessarily predefined.
Given the different levels of training supervision, OVSeg methods are categorized as mask-supervised~\cite{lseg, segclip, catseg, openseg, odise, groundeddiff}, image-text pair supervised~\cite{vilseg, groupvit, segclip, viewco, mixreorg, ovsegmentor, tcl} (hence weakly-supervised), or training-free~\cite{zeroseg, sclip, ovdiff, reco, diffsegmenter}.
Among them, this work corresponds to the last one: OVSeg with no additional training by leveraging pretrained neural networks.
For example, one of the training-free methods, SCLIP~\cite{sclip}, replaces the attention layer of the query and the key cross-similarity into the self-similarity of each embeddings such that it attends on the homogeneous contexts in local regions. 
Our work is distinguished from the existing work~\cite{sclip, clipdinoiser, maskclip} which directly classifies each image pixel with pixel-text similarity without comprehending the objects.
On the other hand, our method first identifies the multiple objects in the absence of the text information and later assigns the text entities to the identified objects in a \textit{lazy visual grounding} fashion.
Our training-free method is also distinguished from the trainable OVSeg methods~\cite{ghiasi2022scaling, xu2022simple, liang2023open, ding2022open} that train MaskFormer segmentation models~\cite{maskformer, mask2former} to obtain class-agnostic masks followed by grounding.



\begin{figure*}[t!]
	\centering
	\small
    \includegraphics[width=\linewidth]{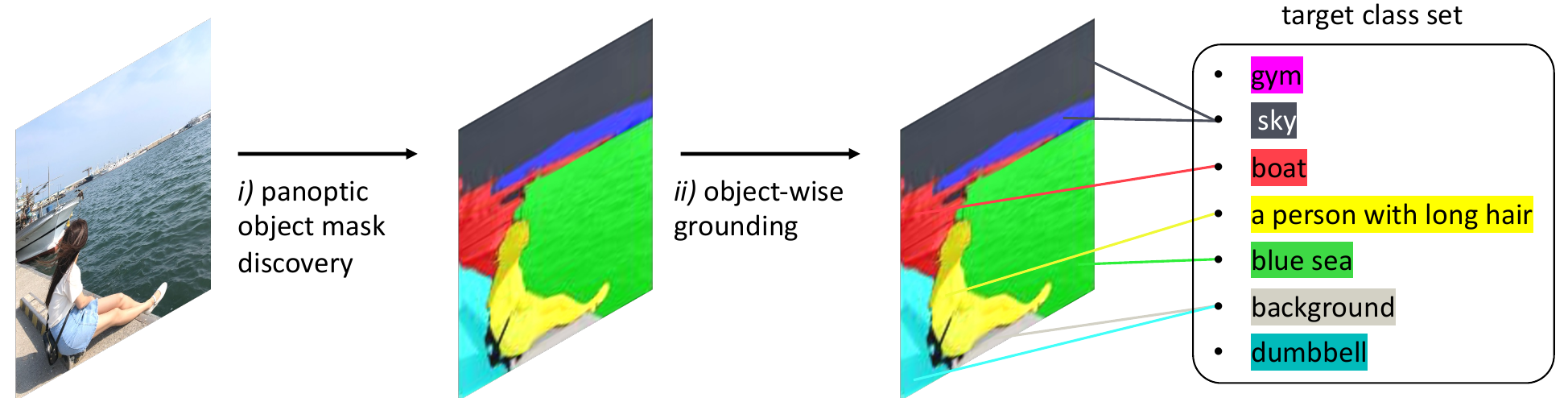}
 \caption{\textbf{Two stages of Lazy Visual Grounding (LaVG).} Given an image, LaVG first discovers existing object masks without the text information (panoptic cut) and then later assigns the class in text descriptions to each object with cross-modal similarity (object grounding).
 }
\label{fig:overview}
\end{figure*}

\section{Lazy Visual Grounding}
We introduce \textit{\textbf{La}zy \textbf{V}isual \textbf{G}\seqsplit{rounding} (LaVG)}.
LaVG consists of the two consecutive stages: 
panoptic object mask discovery (\Secref{sec:discovery}) and object grounding with text (\Secref{sec:text}).
The key theme is \textit{late text interaction};
the visual object discovery stage is a vision-only process and does not involve text interaction.
\Figref{fig:overview} briefly visualizes the process.

\subsection{Panoptic Object Mask Discovery with Normalized Cut}
\label{sec:discovery}

\paragraph{\textbf{Preliminary: Normalized cut.}}
Normalized cut~\cite{normalizedcut} tackles the image segmentation problem with graph partitioning. 
Given an undirected weighted graph of $(\mathcal{V}, \mathcal{E})$ with $N$ nodes, Normalized cut finds the optimal partition of the graph nodes into two disjoint sets of nodes $\mathcal{A}$ and $\mathcal{B}$ such that the removed weights of the edges between $\mathcal{A}$ and $\mathcal{B}$ be the minimum~\cite{wu1993optimal}.
The sum of the lost weights between the bipartitioned subgraphs is defined as \textit{cut}:
\begin{align}
    \text{cut}(\mathcal{A}, \mathcal{B}) = \sum_{u \in \mathcal{A}, v \in \mathcal{B}} w(u, v).
\end{align}
In the image segmentation perspective, the subgraphs $\mathcal{A}$ and $\mathcal{B}$ correspond to two image segments that cover an image.
The bipartition should avoid \textit{bad partitioning} such as a tiny fraction and the rest of the mass, which still satisfies the definition of the minimum cut.
Thus, a more preferable cut is a bipartition that balances the sizes of subgraphs, where each subgraph is properly associated with the entire graph.
Formally,
\begin{align}
    \text{Ncut}(\mathcal{A}, \mathcal{B}) &= \frac{\text{cut}(\mathcal{A}, \mathcal{B})}{\text{assoc}(\mathcal{A}, \mathcal{V})} + \frac{\text{cut}(\mathcal{A}, \mathcal{B})}{\text{assoc}(\mathcal{B}, \mathcal{V})}, \label{eq:ncut} \\
    \text{where}\;\; \text{assoc}(\mathcal{A}, \mathcal{V}) &= \sum_{u \in \mathcal{A}, t \in \mathcal{V}} w(u, t).
\end{align}
The association factor normalizes the cuts by the connected proportion to the entire graph.
Through the systematic derivation, the work of Normalized cut~\cite{normalizedcut} shows that minimizing \Eqref{eq:ncut} is equivalent to solving an eigenvector system.
Let us denote that $\bm{W} \in \Real^{N \times N}$ is a symmetric matrix of the non-negative edge weights, and $\bm{D}$ is a diagonal matrix of which diagonal elements correspond to the row-wise sum of $\bm{W}$. 
Then the solution of Normalized cut is obtained by solving the following generalized eigenvalue system:
\begin{align}
  (\bm{D} - \bm{W})\bm{x} = \bm{\lambda} \bm{D} \bm{x} \label{eq:eigensystem}
\end{align}
and obtaining the eigenvector corresponding to the second-smallest eigenvalue\footnote{The smallest eigenvalue of $\bm{D} - \bm{W}$, \ie, a Laplacian matrix, is always zero. The second-smallest eigenvalue is known as the algebraic connectivity, or Fiedler eigenvalue, and reflects how well connected the graph is.}.
Let us denote the eigenvector corresponding to the second-smallest eigenvalue as $\bm{z} \in \Real^{N}$.
The set of nodes $\mathcal{V} = \{ v_1, \cdots, v_N \}$ is divided into $\mathcal{A}$ and $\mathcal{B}$ by the following bipartition rule:
\begin{align}
  v_n \in \mathcal{A} & \quad \text{if}\ \bm{z}_n > \bar{\bm{z}}, \\
  v_n \in \mathcal{B} & \quad \text{otherwise,} \label{eq:bipartition}
\end{align}
where $\bar{\bm{z}}$ means the average of the vector elements of $\bm{z}$.
For the detailed derivation, please refer to the original paper~\cite{normalizedcut}.

\begin{figure*}[t!]
	\centering
	\small
    \includegraphics[width=0.9\linewidth]{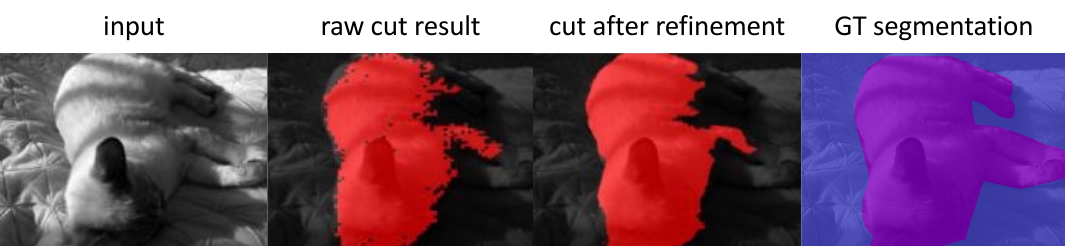}
 \caption{The result of Normalized cut~\cite{normalizedcut} before and after refinement.
 }
\label{fig:eigen}
\vspace{-4mm}
\end{figure*}

\paragraph{\textbf{Panoptic cut.}}
The unsupervised object mask discovery step inherits the idea of the recent object discovery work~\cite{tokencut, cutler, locate} based on Normalized cut and is adapted to semantic segmentation.
Specifically, the recent work~\cite{tokencut, cutler, locate} presents to leverage the DINO ViT~\cite{dino, vit} feature map as an input feature to Normalized cut.
As the task of interest is to identify objects and also assign a semantic class for all image pixels, \ie, panoptic segmentation, we call this unsupervised object mask discovery step as \textit{Panoptic cut}.

Given an input image, the image is fed to a pre-trained self-supervised DINO ViT to obtain the image feature map.
Following the existing work~\cite{amir2021deep, cst, ziegler2022self}, we use the intermediate byproduct during the feature feed-forward process as the dense feature map as it is empirically shown to highlight the salient foreground objects.
In the last layer of the ViT, the patch token similarity between the key embedding produces an affinity map:
\begin{equation}
 \bm{W}_{x {x'}}\!=\!\frac{\bK_x^{\top} \bK_{x'}}{\|\bK_x\|_2 \|\bK_{x'}\|_2},
\end{equation}
where $x, x'$ indicates positional indices.
In the graph partitioning perspective, the set of patch key embeddings correspond to nodes, and their cosine similarity values correspond to the edge weights.
Such definition of graph is applied to the Normalized cut system in Equation~\ref{eq:eigensystem}.

The solved system yields two sets of nodes, $\mathcal{A}$ and $\mathcal{B}$.
We determine the foreground between them by following the few heuristics from MaskCut~\cite{cutler}.
$\mathcal{A}$ is foreground if a) $\mathcal{A}$ contains the maximum absolute value in the second smallest eigenvector $\bm{z}$ and b) $\mathcal{A}$ is covering less than the two corners of the matrix.
Once we determine the foreground between $\mathcal{A}$ and $\mathcal{B}$, then $\mathcal{A}$ is discovered as an object, and we iterate bipartitioning the rest of the background node set until i) there are less than five nodes left or ii) the iteration count reaches the preset maximum\footnote{In implementation, the condition i) is rarely met. We empirically observe that DINO features exhibit small activation values on non-salient regions. Consequently, Panoptic cut returns bad bipartitions, which is ignored during the mask refinement process.
}.
After the iteration halts, the undiscovered region is marked as background.
We often observe noisy object boundaries or the salt and pepper artifacts such as in \Figref{fig:eigen}.
We thus fill the holes and refine the output bipartition masks with DenseCRF~\cite{densecrf},
before the masks are resized to the original image size.
We note that the process introduced so far uses neither human annotation nor text entity.

Panoptic cut distinguishes from the unsupervised object discovery models~\cite{cutler, locate, tokencut} in that the background region is also semantically segmented.
For example, the background of sky and ceiling are not often perceived as objects, but those pixels can be partitioned with different semantics.

\subsection{Object Grounding with Class Descriptions in Text}
\label{sec:text}

\paragraph{\textbf{Object mask grounding with cross-modal similarity.}}
We adopt an image-text multi-modal encoder backbone to assign a text entity to each discovered mask by cross-modal similarity matching.
We use the patch embedding of the SCLIP~\cite{sclip} backbone which roughly aligns the local image context with each patch embedding.
Given a set of $Y$ text classes, the text class descriptions are fed to the CLIP text encoder to obtain the set of text embeddings: $\{\vc_y\}_{y=1}^{Y}$.
Note that some benchmarks contain the background class.
In these cases, following \cite{sclip}, we feed the background query text descriptions, \eg, wall, sky, \etc, to avoid the ambiguous semantic of the text ``background''.
The input image is fed to the SCLIP image encoder which returns the feature map: $\{\vf_x | \, x \in [1, H] \times [1, W] \}$, where $x$ indicates a position.
The image features that fall into an object mask region $\mathcal{M}$ are averaged to construct the \textit{object prototype}~\cite{protonet, wang2019panet}.
The class distribution is obtained by the cosine similarity of the object prototype and the text embedding:
\begin{equation}
    p(\vf_{x \in \mathcal{M}}|y) \propto \frac{1}{|\mathcal{M}|} \sum_{x' \in \mathcal{M}} \vf_{x'}^{\top} \vc_y \label{eq:objectprototype}
\end{equation}
to assign the text class on the object, \ie, object-wise grounding.
     

\paragraph{\textbf{Segmentation prediction.}}
The class logit map is resized to the input image size by bilinear interpolation.
When the background query set is used, the corresponding logits are merged as background.
Each object mask is predicted with the highest cross-modal similarity class index.


\begin{table*}[t!]
    \centering
    \captionof{table}{Benchmarks for open-vocabulary semantic segmentation}    \label{table:datasets}
    \tabcolsep=0.05cm
    \scalebox{0.9}{
    \begin{tabular}{lccccccc}
    \toprule
    \multirow{2}{*}{} & \multicolumn{3}{c}{\it With a background category} & \multicolumn{4}{c}{\it Without background category} \\\cmidrule(lr){2-4}\cmidrule(lr){5-8}
    & VOC21 & Context60 & COCO-Object & VOC20 & Context59 & ADE20K & COCO-Stuff \\\midrule
    number of classes & 21 & 60 & 81  & 20 & 59 &  150 &  171 \\ \bottomrule
    \end{tabular}
    }
    \vspace{-5mm}
\end{table*}

\section{Experiments}

\subsection{Experimental Setup}

\begin{figure*}[t]
	\centering
	\small
    \includegraphics[width=\linewidth]{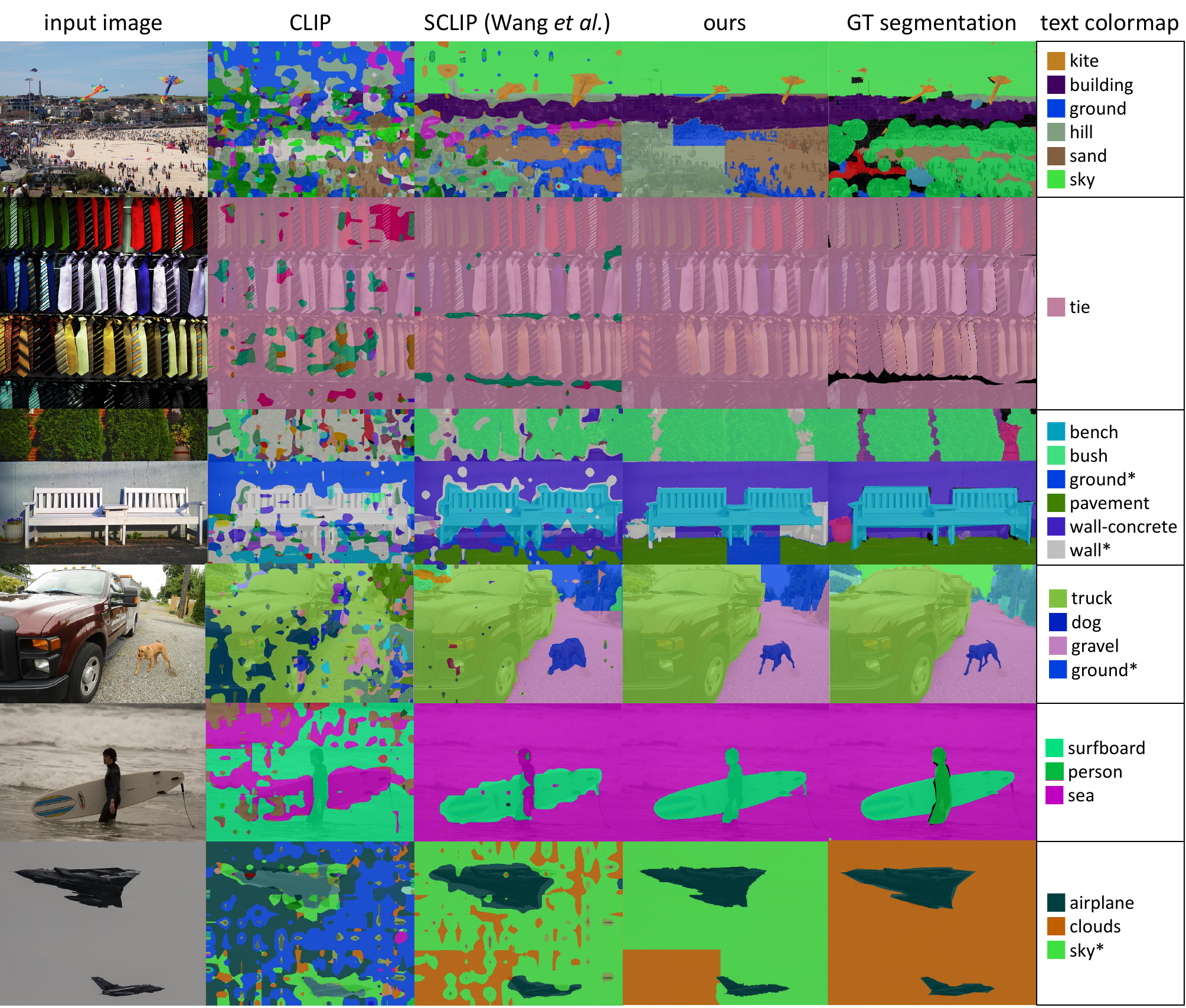}
 \captionof{figure}{Qualitative comparison of our method and baselines on COCO-stuff 164K.  
 The * mark on the text set denotes the false positive class prediction of our model.
 }
\label{fig:qual}
\end{figure*}

\paragraph{\textbf{Implementation details.}}
For the Panoptic cut input source, we use the feature map from DINO ViT-B/8~\cite{dino, vit} trained on ImageNet1K~\cite{russakovsky2015imagenet} without labels.
To solve the generalized eigenvector system in Equation~\ref{eq:eigensystem}, we use the PyTorch~\cite{pytorch} builtin function $\mathtt{lobpcg}$ which is compatible with GPUs.
The maximum iteration for Panoptic cut is set to 16.
We also use the pre-trained CLIP ViT-B/16~\cite{clip} for the image/text feature extractor.
For object grounding with CLIP, we adopt the sliding-window segmentation prediction scheme on high-resolution images, which is shown effective from the previous work~\cite{groupvit, tcl, sclip}.
We upsample the image as well as the Panoptic cut object masks such that the shorter size of the image becomes 336.
The sliding window and the stride are set to $224\times224$ and $112\times112$, respectively.
Then each crop is fed to the image encoder which returns the pixel-level logit map on each cropped window, followed by averaging the logits.
The codebase is adopted from SCLIP~\cite{sclip} which is based on \texttt{mmseg}~\cite{mmseg2020}.
An Nvidia RTX 3090 GPU is used for all experiments.

\paragraph{\textbf{Datasets and benchmarks.}}
We evaluate our method on five publicly available open-vocabulary semantic segmentation datasets: Pascal VOC 2012~\cite{pascal}, Pascal Context~\cite{pascalcontext}, ADE20K~\cite{ade20k}, COCO-Stuff and COCO-Object~\cite{coco} plus two variants with the background queries.
Background queries such as wall, sky, \etc are 
added on Pascal VOC and Pascal Context, which are indicated as VOC21 and Context60, respectively.
We use the background query set from SCLIP~\cite{sclip} and do not engineer it further.
Table~\ref{table:datasets} shows the configuration of the dataset classes.
For evaluation, we use the standard mean IoU metric: $\mathrm{mIoU} = \frac{1}{Y}\sum_y \mathrm{IoU}_y$, where $\mathrm{IoU}_y$ denotes an IoU value of $y$-th
class.

\begin{figure*}[t!]
	\centering
	\small
    \includegraphics[width=\linewidth]{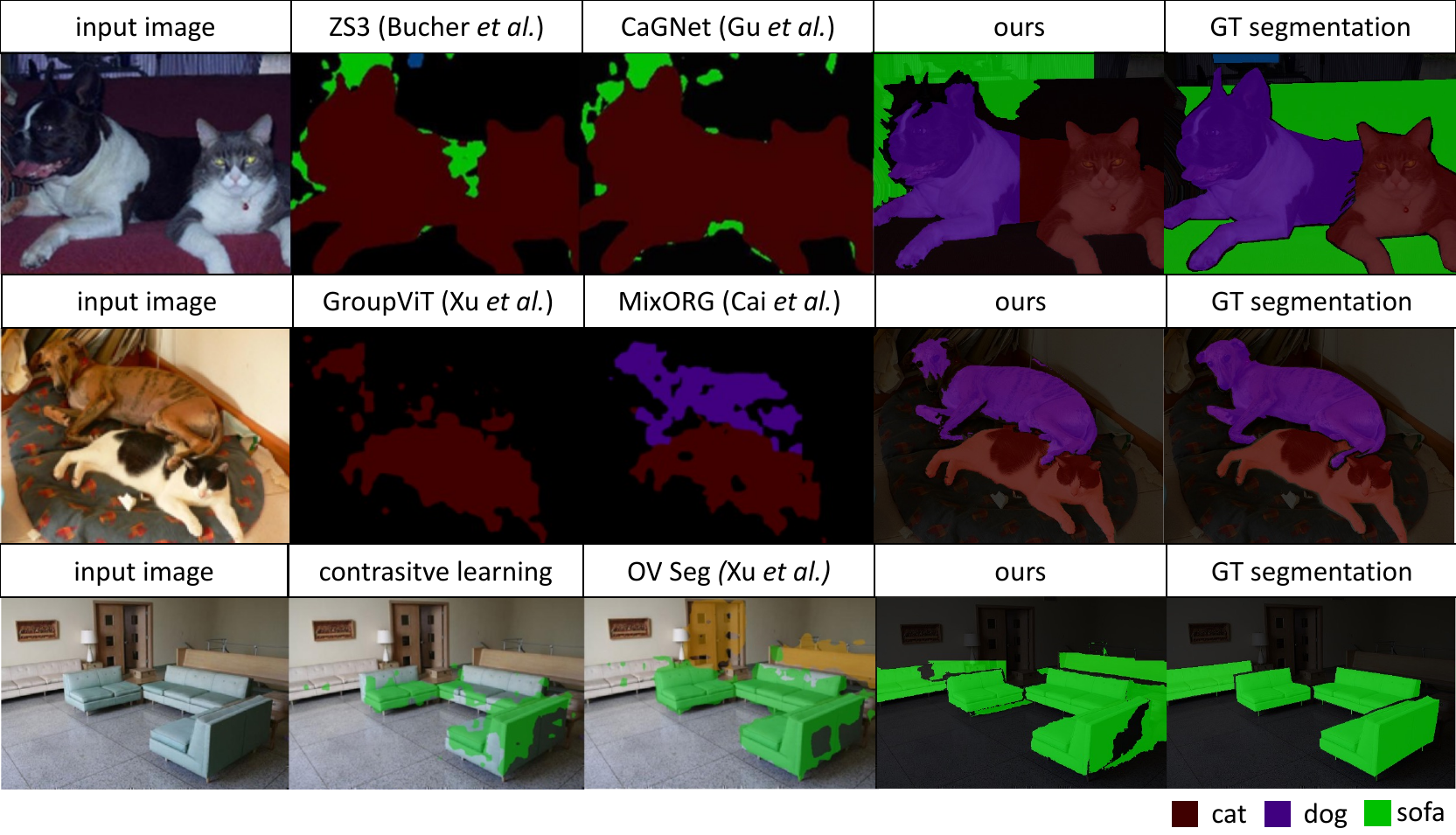}
 \caption{Qualitative comparison of our method and others. 
 The first three columns are directly copied from the paper in the third columns~\cite{z3net, cagnet, groupvit, mixreorg, ovsegmentor}.
 }
\label{fig:otherqual}
\vspace{-3mm}
\end{figure*}

\begin{table*}[t]
    \centering
    \caption{Performance (mIoU, \%) comparison of the state-of-the-art open-vocabulary segmentation methods on seven public benchmarks.
    The models with $\dagger$ do not use CLIP. 
    The models with $\checkmark$ involve additional training on top of the pretrained backbone.
    The numerical results~\cite{clip, maskclip, shin2022reco, tcl, ovsegmentor, segclip} are adopted from SCLIP~\cite{sclip}.
    }
    \tabcolsep=0.04cm
    \centering
    \scalebox{0.91}{
    \begin{tabular}{lcccccccc}
    \toprule
    \multirow{2}{*}{\textbf{Method}} & & \multicolumn{3}{c}{\it With a background category} & \multicolumn{4}{c}{\it Without background category} \\\cmidrule(lr){3-5}\cmidrule(lr){6-9}
    & train & VOC21 & Context60 & COCO-Obj. & VOC20 & Context59 & ADE & COCO-Stf.\\\midrule
    CLIP~\cite{clip} & & 18.8 & 9.9 & 8.1 & 49.4  & 11.1 & 3.1 & 5.7  \\
    ZeroSeg~\cite{zeroseg} & & 40.8 & 20.4 & 20.2 & - & - & -& -\\ 
    MaskCLIP~\cite{maskclip} & & 43.4 & 23.2 & 20.6 & 74.9 & 26.4 & 11.9 & 16.7 \\ 
    ReCo~\cite{shin2022reco} & & 25.1 & 19.9 & 15.7 & 57.7 & 22.3 & 11.2 & 14.8 \\ 
    CLIPpy~\cite{clippy} & & 52.2 & - & 32.0 & - & -  & 13.5 & -  \\ 
    DiffSegmenter$^{\dagger}$~\cite{diffsegmenter} & & 60.1 & 27.5 & \bf 37.9 & - & - & - & -  \\ 
    SCLIP~\cite{sclip} & & 59.1 & 30.4 & 30.5 & 80.4 & 34.2 & \bf 16.1 & 22.4  \\  
    ViL-Seg~\cite{vilseg} & \checkmark & 34.4 & 16.3 & 16.4 & - & - & 11.9 & - \\ 
    ViewCo~\cite{viewco} & \checkmark & 52.4 & 23.0 & 23.5 & -  & - & - & -\\ 
    MixReorg~\cite{mixreorg} & \checkmark & 47.9 & 23.9 & - & - & - & - & -\\ 
    GroupViT$^{\dagger}$~\cite{groupvit} & \checkmark & 52.3 & 18.7 & 27.5 & 79.7 & 23.4 & 10.4 & 15.3 \\  
    TCL~\cite{tcl}           & \checkmark & 51.2 & 24.3 & 30.4 & 77.5 & 30.3 & 14.9 & 19.6 \\ 
    OVSegmentor$^{\dagger}$~\cite{ovsegmentor} & \checkmark & 53.8 & 20.4 & 25.1 &  - & - &  5.6 & - \\ 
    SegCLIP~\cite{segclip}   & \checkmark & 52.6 & 24.7 & 26.5    & -    & -    & -        \\  
    \rowcolor{gray!10}
    \multicolumn{2}{l}{LaVG (ours) } & \bf 62.1 & \bf 31.6 &  34.2 & \bf 82.5 & \bf 34.7 & 15.8 & \bf 23.2 \\ 
    \bottomrule
    \end{tabular}
    }
    \label{table:main}
\end{table*}

\paragraph{\textbf{Baseline methods.}}
Vanilla CLIP is the most na\"ive baseline.
The set of image patch embeddings and the set of text embeddings are obtained from the pre-trained CLIP image and text encoder backbone, respectively.
Cosine similarity between the two sets of embeddings is computed to obtain the cross-modal pixel-wise text similarity, which is directly translated as semantic segmentation prediction.
SCLIP~\cite{sclip} is the primary baseline.
SCLIP modifies the intermediate function of the attention layers in the pre-trained CLIP ViTs such that the final image patch embedding better represents the corresponding image patch semantics.

\subsection{Experimental Results}

\paragraph{\textbf{Qualitative results.}}
We demonstrate the qualitative examples of the model outputs in \Figref{fig:qual}.
Our model demonstrates advanced segmentation visuals compared to our baselines.
Note that CLIP and SCLIP~\cite{sclip} are pixel-wise dense prediction methods.
When segmentation prediction operates at pixel level, it does not necessarily comprehend the object shapes, resulting in noisy segmentation masks of objects.
Ours, however, proceeds object mask discovery stage and returns object-level masks.
As seen, our model predicts sharper and cleaner delineations that accurately capture semantic object boundaries, \eg, our model segments out the clear shape of the kites in the sky (first image).
Other nice properties that our model often exhibits include capturing tiny kites (first image), segmentation with little spurious correlation (second image), and repeated chairs (third image).  

We also compare various other OVSeg models~\cite{z3net, cagnet, groupvit, mixreorg, ovsegmentor} in \Figref{fig:otherqual}.
We copy and paste the visualization on their original papers and present our model's prediction.
Thus the results are not much cherry-picked as the choice of visualized images is taken from the other work.
Most previous work often presents visualized prediction favorably on Pascal VOC, most of which contain the objects that stand out.
Unlike the examples of the other work with incorrect class predictions (the dog in the first image), holes in predicted masks (the second image), and misaligned contour (third example),
our model presents much cleaner mask prediction and better class prediction.


\paragraph{\textbf{Quantitative comparison with state-of-the-art models.}}
\Tableref{table:main} compares the existing OVSeg methods on seven public benchmarks.
The performance values of MaskCLIP~\cite{maskclip}, ReCo~\cite{reco}, and TCL~\cite{tcl} are brought from the reimplementation of SCLIP, which are either higher or the same values as the ones in the original papers.
The performance values of the rest of work~\cite{vilseg, zeroseg, viewco, mixreorg, groupvit, clippy, ovsegmentor, segclip, diffsegmenter} are from their original performance reports.
Our model outperforms the previous state-of-the-art models.
Note that our model outperforms not only the training-free models~\cite{sclip, clippy, zeroseg, maskclip, reco} but also the weakly-supervised learning models~\cite{vilseg, groupvit, segclip, viewco, mixreorg, ovsegmentor, tcl} that exploits image-text paired datasets, \eg, CC3M~\cite{cc3m}, YFCC~\cite{yfcc}, or COCO~\cite{coco} from the 3.4 M to 26 M dataset scale.

\begin{table*}[t!]
    \centering
    \caption{Comparison with different object mask discovery models}
    \tabcolsep=0.04cm
    \scalebox{0.92}{
    \begin{tabular}{lcccccc}
    \toprule
    & VOC21 & Context60 & COCO-Obj. & VOC20 & Context59  & COCO-Stf. \\\midrule
    Normalized cut~\cite{normalizedcut} & 47.6 & 26.3 & 28.2 & 78.1 & 29.3 & 20.0 \\ 
    CutLER-CascadedRCNN~\cite{cutler} & 57.3 & 29.3 & 33.2 & 79.6 & 32.2 & 22.3 \\ 
    Panoptic cut (4 iterations) & 61.1 & 30.3 & 33.7 & 81.3 & 33.4 & 22.6 \\
        \rowcolor{gray!10}
    Panoptic cut (16 iterations) & \bf 62.1 &  31.6 & \bf 34.2 & 82.5 & 34.7 & 23.2 \\
    Panoptic cut (32 iterations) & 61.4 & \bf 31.7 & \bf 34.2 & \bf 82.6 & \bf 34.8 & \bf 23.3 \\ 
    \gray{Ground-truth object masks} & \gray{68.3} & \gray{37.1} & \gray{44.1} & \gray{87.7} & \gray{40.2} & \gray{27.6} \\ \bottomrule
    \end{tabular}
    }
    \label{table:od}
\end{table*}

\subsection{Experimental Analyses}
We ablate the components in our method and see their effectiveness.
In tables, our method is colored in gray.

\begin{table*}[t!]
    \centering
    \caption{Comparison with different grounding schemes. $\mathcal{O}$ denotes object mask region.}
    \tabcolsep=0.04cm
    \scalebox{0.95}{
    \begin{tabular}{lccccccc}
    \toprule
    & VOC21 & Context60 & COCO-Object & VOC20 & Context59  & COCO-Stuff \\\midrule
    $\mathcal{O}$-masked attention & 54.6 & 26.9 & 27.6 & 61.7 & 28.2 & 18.3 \\ 
    Gaussian blur on non-$\mathcal{O}$ & 59.8 & 30.9 & 31.5 & 71.6 & 33.4 & 21.5 \\ 
    \rowcolor{gray!10}
    $\mathcal{O}$-averaged prototype & \bf 62.1 & \bf 31.6 & \bf 34.2 & \bf 82.5 & \bf 34.7 & \bf 23.2  \\ \bottomrule
    \end{tabular}
    }
    \label{table:textgrounding}
\end{table*}

\paragraph{\textbf{Comparison with other object mask discovery models.}}
To verify the strength of Panoptic cut in segmentation, we compare Panoptic cut with other object discovery models and compare them in \Tableref{table:od}.
Among the modern object discovery neural networks, CutLER~\cite{cutler} is compared.
CutLER first produces object mask proposals with a derivative of Normalized cut (MaskCut) and uses the proposals as self-supervision to train an unsupervised instance segmentation \& object detection architecture, \eg, Cascaded-RCNN~\cite{cascadercnn}.
To adapt CutLER to semantic segmentation, the undetected pixels are considered as a single region and averaged at the feature level as background.
The na\"ive adoption of the state-of-the-art object discovery method does not directly translate to high open-vocabulary segmentation performance.
We observe that CutLER tends to detect fine-grained parts of foreground objects \textit{while leaving the background under-segmented}, \eg, undetected sky and sea grouped together, probably because the training objective is to detect foreground objects.
Although this model achieves promising performance that beats most of the methods in \Tableref{table:main} but is lower than Panoptic cut.
Panoptic cut produces proper semantic segmentation of background, plus, it requires  no additional training.
We believe that Panoptic cut is a more suitable choice than instance detection or salient detection methods~\cite{hsu2018unsupervised} for semantic segmentation, where all the present pixels should be classified for segmentation.
Panoptic cut iterates the preset number of iterations, where we gain from 0.5 \% to 1.3 \% point mIoU gain from maximum of 4 to 16 iterations.


\paragraph{\textbf{Comparison with different grounding schemes.}}
We explore different grounding schemes and compare them on \Tableref{table:textgrounding}.
Specifically, three object embedding methods are compared, all based on the SCLIP feature.
First, object-masked $\mathcal{O}$-masked attention~\cite{ifsl, mask2former, bert} forces the forward path of ViTs to attend only to the designated region.
We zero out the non-object region on the input image and set a large negative value on the non-object pixel patch embedding before the softmax in the attention in the ViT backbone.
Recent work~\cite{shtedritski2023does} explores a way to explicitly draw attention on the input image level in a human-friendly manner.
We adopt their idea of applying Gaussian blur on the non-object regions on the input image and taking the global image representation as the local feature.
Note that these two schemes require multiple feed-forward passes with the same image with different attention masks and different Gaussian blurred regions.
Our method (the last line) is the simplest: averaging the region patch embedding (Eq.~\ref{eq:objectprototype}).
The object-averaged prototype shows superior performance in terms of both end-task accuracy and complexity.

\paragraph{\textbf{Computational complexity.}}
\Tableref{table:complexity} compares the computational cost of LaVG with a training-free model~\cite{diffsegmenter} and trained models~\cite{xu2022simple, catseg}.
The outstanding advantage of training-free models is the computational efficiency from the absence of time, data, and computation burden in additional training.
Compared to another training-free ovseg method~\cite{diffsegmenter} based on diffusion models~\cite{stabledifussion} which leverages image/caption generation models with heavy parameters, we use much more lightweight feature embedding models.
However, our iterative process of panoptic cut introduces a longer inference time. 



\begin{table}[t]
\centering
\caption{Comparison of computational cost measured with an Nvidia RTX 3090. The columns with $\dagger$ are taken from \cite{catseg}. \label{table:complexity}}
\tabcolsep=0.05cm
\scalebox{0.98}{
\begin{tabular}{l|ccccc}
    \toprule
    methods & ZSSeg~\cite{xu2022simple}$^\dagger$ & CATseg~\cite{catseg}$^\dagger$ & DiffSegmenter~\cite{diffsegmenter} & LaVG \\
    \midrule
    \# learnable params. (M) & 102.8  & 70.3 & 0 &  0\\
    \# total params. (M) & 530.8 & 433.7 & 1453 & 198 \\
    training time (min) & 958.5 & 875.5 & 0 & 0 \\
    inference time (s) & 2.73 & 0.54 & 0.59 & 12.59 \\
    \bottomrule
\end{tabular}
}
\end{table}





\subsection{Limitations of our approach}
We outline four characteristic limitations of our method.
The intrinsic problem of our method is the extensive computation time and memory consumption induced by leveraging the graph-based partitioning algorithm.
Normalized cut generally consumes excessive use of memory due to the use of $N \times N$ size affinity matrix.
For example, running Normalized cut on a 512 $\times$ 512 RGB image directly leads to the out-of-memory error.
However, using a well-abstracted downsampled feature map of the DINO attention greatly reduces the memory consumption and smoothly executes on a single regular GPU device, \eg, Nvidia RTX 3090.
Also, our approach is inference-only thus includes no training time, which is still more computationally efficient than training models.

In terms of the end-task performance, we witness the remarkable qualitative improvement as seen in Figures~\ref{fig:qual}-\ref{fig:otherqual}, but it does not directly enhance mIoU values substantially.
This is because the evaluation metric is pixel-level classification rather than favoring aligned edges or clean shapes of objects.
In addition, the proposed method occasionally misclassifies objects which also negatively affects the mIoU performance.

Another limitation is qualitatively observed in the last example of \Figref{fig:qual}.
Due to the sliding-window segmentation inference technique, the cropped region suffers the part-whole ambiguity issue and sometimes predicts a wrong class on the region. 
This side effect is often observed on textureless regions, 
\eg, the model confuses the sky and cloud in the last example of \Figref{fig:qual}.
However, the recognition of the image as a whole might remedy the part-whole ambiguity, but the accuracy of the small-resolution images drops at the same time.

Lastly, out method implemented with CLIP is potentially inappropriate for closed-vocabulary segmentation on specialized domains such as medical or industrial vision.
For such problems on specific domains, however, the concept of lazy grounding can be used with domain-specific encoders.


\section{Conclusion}
\label{sec:conclusion}
We have proposed Lazy Visual Grounding, a late-interaction semantic segmentation approach, for open-vocabulary semantic segmentation.
Our method first proceeds unsupervised object mask discovery by graph partitioning followed by grounding on the discovered object masks.
This simple and training-free method exceeds the performance of the caption-based weakly-supervised learning models.
Given the observation that the classic graph partitioning method surpasses deep-learning models,
we claim that semantic \seqsplit{pixel-grouping} is crucial for semantic segmentation and should be prioritized especially when segmentation supervision is not available.
Last but not least, as our method produces object-level segmentation outputs with crisp \textit{object} boundaries, our method is particularly more suitable for grounding tasks requiring object-level identification with precise object boundaries.
This work may open the door for training-free open-vocabulary \textit{instance} segmentation, which has not been actively studied, yet.

\paragraph{\textbf{Acknowledgements.}} 
This work was supported by Samsung Electronics \seqsplit{(IO201208-07822-01)}, the NRF grant (NRF-2021R1A2C3012728 ($45\%$)), and the IITP grants (RS-2022-II220959: Few-Shot Learning of Causal Inference in Vision and Language for Decision Making ($50\%$), RS-2019-II191906:
AI Graduate School Program at POSTECH ($5\%$)) funded by Ministry of Science and ICT, Korea.
We also thank Sua Choi for her helpful discussion.

\title{
In Defense of Lazy Visual Grounding
\\ for Open-Vocabulary Semantic Segmentation
\\ -- \textit{Supplementary Materials} --}







\section{Appendix}
In this appendix, we provide additional experimental details and results of our method.

\subsection{Terminology Reference to the ``Lazy'' Action}
The notion of \textit{lazy} action is broadly used 
in the context of computer software design.
Lazy action refers to the generic design choice where an action is defined but defers launching until immediately required.
For instance, in \textbf{web programming}~\cite{celik2016build, green2013angularjs}, the lazy loading/fetching strategy defers the actual instantiation of a web object until it is accessed, \eg, clicked.
In the SQL \textbf{database management system}~\cite{faleiro2014lazy, goncalves2013java}, lazy connecting~\cite{sqlalchemylazyconnecting} refrains from establishing connection to the database until it is explicitly required to execute a specific operation on the database.
In the virtual memory design of \textbf{operating systems}, lazy swapping~\cite{oslazyswapping}, also known as demand paging, selectively loads portions of a program into RAM.
It defers loading the remaining portions until the already loaded part specifically accesses them.
Conversely, \textit{eager} actions refer to the opposite concept;
it immediately executes actions such as data instantiation or memory loading \textit{right away} without waiting for designated requirements.
Lazy action effectively reduces the operation time especially when it is mostly unnecessary or computationally redundant to launch actions eagerly.

In our context of image semantic segmentation, our model discovers the objects first, and then visual grounding takes place afterward.
This lazy grounding strategy greatly reduces the spurious pixel-wise noise, which the eager grounding methods~\cite{sclip, zeroseg, tcl, maskclip} often suffer during pixel-level grounding, and achieves the remarkable segmentation quality improvement.

\begin{table*}[t!]
    \centering
    \caption{Comparison of open-vocabulary semantic segmentation models. For unsupervised object mask discovery, DINO or SCLIP is used for  Panoptic cut (PC) in our model. }
    \label{table:clipdiscovery}
    \tabcolsep=0.04cm
    \scalebox{0.9}{
    \begin{tabular}{lcccccccc}
    \toprule
    \multirow{2}{*}{\textbf{Method}} & additional & \multicolumn{3}{c}{\it With a background category} & \multicolumn{4}{c}{\it Without background category} \\\cmidrule(lr){3-5}\cmidrule(lr){6-9}
    & training & VOC21 & Context60 & COCO-Obj. & VOC20 & Context59 & ADE & COCO-Stf.\\\midrule
    CLIP~\cite{clip} & & 18.8 & 9.9 & 8.1 & 49.4  & 11.1 & 3.1 & 5.7  \\\midrule
    ZeroSeg~\cite{zeroseg} & & 40.8 & 20.4 & 20.2 & - & - & -& -\\ 
    MaskCLIP~\cite{maskclip} & & 43.4 & 23.2 & 20.6 & 74.9 & 26.4 & 11.9 & 16.7 \\ 
    ReCo~\cite{shin2022reco} & & 25.1 & 19.9 & 15.7 & 57.7 & 22.3 & 11.2 & 14.8 \\ 
    CLIPpy~\cite{clippy} & & 52.2 & - & 32.0 & - & -  & 13.5 & -  \\ 
    SCLIP~\cite{sclip} & & 59.1 & 30.4 & 30.5 & 80.4 & 34.2 & \bf 16.1 & 22.4  \\  
    ViL-Seg~\cite{vilseg} & \checkmark & 34.4 & 16.3 & 16.4 & - & - & 11.9 & - \\ 
    ViewCo~\cite{viewco} & \checkmark & 52.4 & 23.0 & 23.5 & -  & - & - & -\\ 
    MixReorg~\cite{mixreorg} & \checkmark & 47.9 & 23.9 & - & - & - & - & -\\ 
    GroupViT$^{\dagger}$~\cite{groupvit} & \checkmark & 52.3 & 18.7 & 27.5 & 79.7 & 23.4 & 10.4 & 15.3 \\  
    TCL~\cite{tcl}           & \checkmark & 51.2 & 24.3 & 30.4 & 77.5 & 30.3 & 14.9 & 19.6 \\ 
    OVSegm.$^{\dagger}$~\cite{ovsegmentor} & \checkmark & 53.8 & 20.4 & 25.1 &  - & - &  5.6 & - \\ 
    SegCLIP~\cite{segclip}   & \checkmark & 52.6 & 24.7 & 26.5    & -    & -    & -        \\  
    \rowcolor{gray!10}
    \multicolumn{2}{l}{LaVG \scriptsize{PC with DINO} } & \bf 62.1 & \bf 31.6 & \bf 34.2 & \bf 82.5 & \bf 34.7 & 15.8 & \bf 23.2  \\ 
    \multicolumn{2}{l}{LaVG \scriptsize{PC with SCLIP}} & 47.1 & 29.6 & 23.2 & 80.6 & 32.9  & 12.5  & 18.9\\ 
    \bottomrule
    \end{tabular}
    }
\end{table*}



\subsection{Additional Ablation Studies}
Each following paragraph presents the result of ablating or replacing a single component in our complete method.
Therefore, the results of this section can be fairly compared with the existing methods~\cite{clip, zeroseg, maskclip, shin2022reco, clippy, sclip, vilseg, viewco, mixreorg, groupvit, tcl, ovsegmentor, segclip} found in \Tableref{table:clipdiscovery}.
In Tables~\ref{table:clipdiscovery}-\ref{table:imgsize},
the rows of our approach is colored in gray.

\paragraph{\textbf{Open-vocabulary segmentation of images in the wild.}}
We evaluate the segmentation capability of our model facilitated with the \textit{open vocabulary} on the images in the wild.
The results are demonstrated in Figure 2 in the main paper, where the example images are taken by ourselves with iPhone 14.
Our model exhibits precise semantic segmentation given arbitrary free-form text descriptions such as ``spaghetti carbonara'' or ``a person with long hair''.
We emphasize that these segmentation results are obtained without additional training on the pre-trained DINO and CLIP backbones.

\begin{figure*}[t!]
	\centering
	\small
    \includegraphics[width=\linewidth]{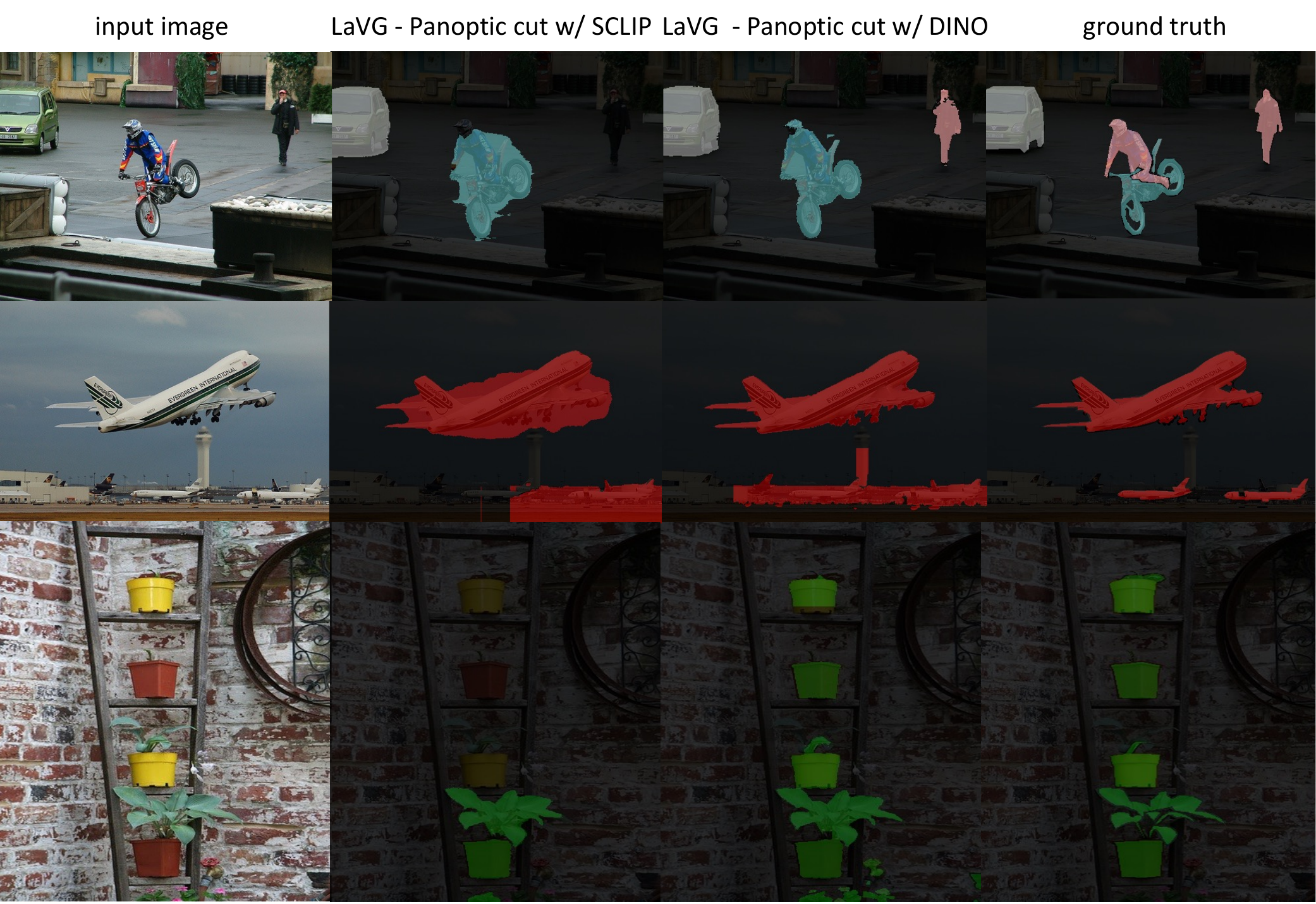}
 \caption{Visual comparison of the backbone used in Panoptic cut for object mask discovery. The CLIP-derived backbone tends to dilate the object edges and occasionally misses small objects.
 We thus choose to implement Panoptic cut on top of DINO.
 }
\label{fig:clipdinodiscovery}
\end{figure*}

\paragraph{\textbf{Effect of the backbone used in Panoptic cut.}}
The Panoptic cut (PC) approach performs on top of the DINO~\cite{dino} backbone for unsupervised object mask discovery.
We attempt to replace it with the SCLIP~\cite{sclip} backbone of the CLIP~\cite{clip} architecture to unify the backbones for the object mask discovery and the visual grounding stages.
\Tableref{table:clipdiscovery} presents the quantitative result, where the CLIP backbone generally underperforms than DINO.
DINO is a vision-only self-supervised image representation model that encourages mapping an image feature and its augmented view closely on the feature space.
CLIP learns to map an image and its alt text closely on a shared vision-and-language embedding space.
Based on this, we suspect that CLIP's image-text matching learning objective is biased toward highlighting the most corresponding region with text description rather than grasping the precise object edge details.
However, DINO is trained with visual input only and focuses more on augmentation-invariant visual clues such as edges and textures.
\Figref{fig:clipdinodiscovery} qualitatively compares the two backbones used in Panoptic cut for object mask discovery,
where the object masks from SCLIP are not as precise as those from DINO.
Specifically, we observe that CLIP-derivative backbones tend to capture blurry object edges (\eg, 1st and 2nd images) and occasionally miss small objects (\eg, 1st and 3rd images).
For this reason, we choose DINO for object mask discovery and SCLIP for visual grounding.

\begin{table*}[t!]
    \centering
    \caption{The effect the sliding-window inference}
    \tabcolsep=0.04cm
    \scalebox{0.92}{
    \begin{tabular}{c|ccccc}
    \toprule
    sliding-window inference & VOC21 & Context60 & COCO-Obj. & VOC20 & Context59   \\\midrule
     & 32.3 & 25.8  & 30.3 & 70.8 &  27.4 \\ 
    \rowcolor{gray!10}
    \checkmark & \bf 62.1 & \bf 31.6 & \bf 34.2 & \bf 82.5 & \bf 34.7 \\ \bottomrule
    \end{tabular}
    }
    \label{table:imgsize}
\end{table*}

\paragraph{\textbf{Effect of sliding-window inference.}}
\Tableref{table:imgsize} presents the ablation study of the sliding window inference technique, which is previously employed in the other work~\cite{ovdiff, tcl, sclip}.
The sliding window technique first resizes the input image such that its shorter side becomes the 336-pixel length and each $224 \times 224$ cropped region is fed to the backbone as explained in Sec.~4.1 in the manuscript.
We observe that the effect of the sliding-window inference fashion is indeed powerful in our lazy grounding model as well.

%
%
\bibliographystyle{splncs04}
\bibliography{dahyun_bib}
\end{document}